\documentclass{article}

    \usepackage[preprint]{neurips_2025}

\usepackage[utf8]{inputenc} % allow utf-8 input
\usepackage[T1]{fontenc}    % use 8-bit T1 fonts
\usepackage{hyperref}       % hyperlinks
\usepackage{url}            % simple URL typesetting
\usepackage{booktabs}       % professional-quality tables
\usepackage{amsfonts}       % blackboard math symbols
\usepackage{nicefrac}       % compact symbols for 1/2, etc.
\usepackage{microtype}      % microtypography
\usepackage{xcolor}         % colors
\usepackage{graphicx}
 \usepackage{amsmath}
 \usepackage{multirow}
\usepackage{multicol}
\title{UP-SLAM: Adaptively Structured Gaussian SLAM with Uncertainty Prediction in Dynamic Environments}

\author{%
  Wancai Zheng
   \\
  \And
   Linlin Ou \\
  \AND
  Jiajie He\\
  \And
  Libo Zhou \\
  \And
  Xinyi Yu
  \And 
   Yan Wei
}

\begin{document}

\maketitle

\begin{figure}[htb]
    \centering
    \includegraphics[scale=0.19]{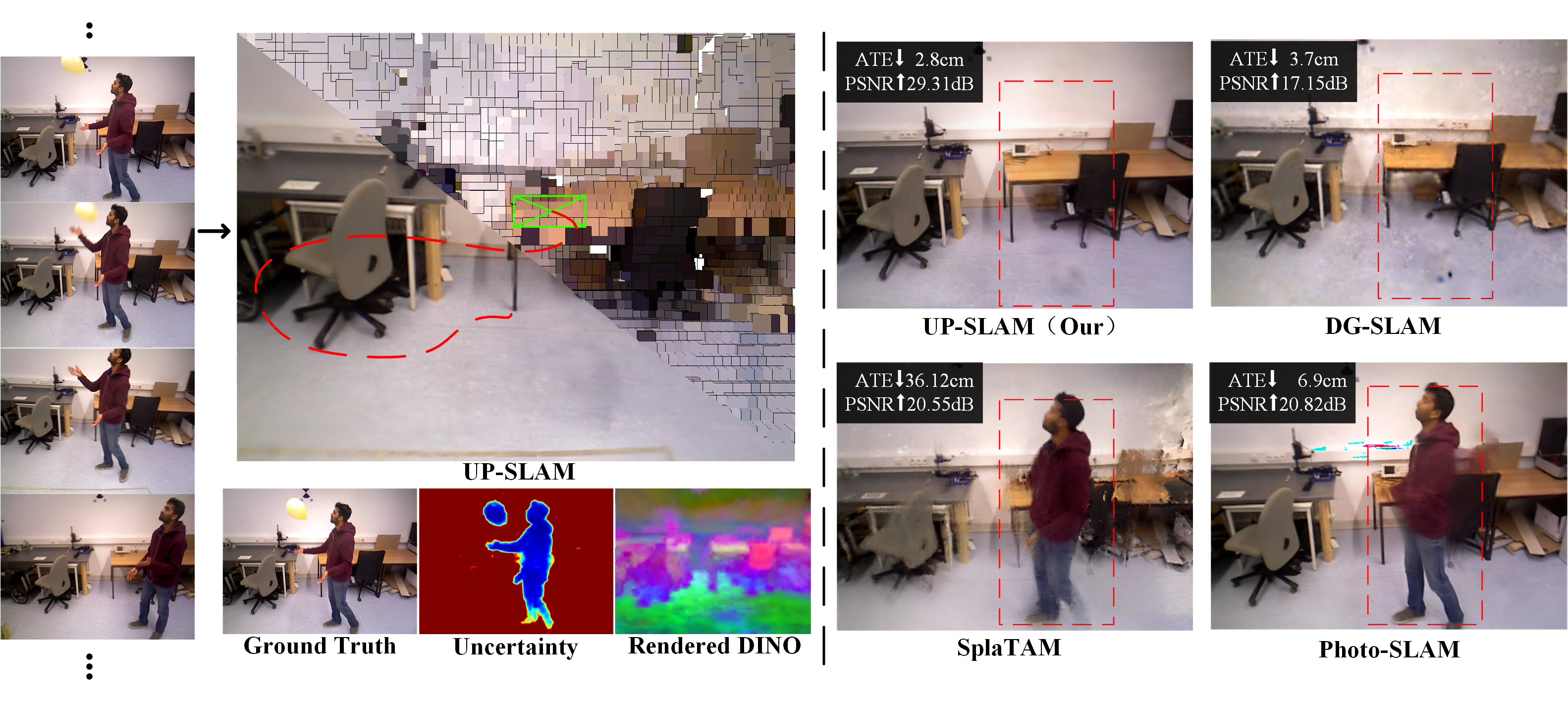}
    \caption{\textbf{UP-SLAM} compresses the 3DGS~\cite{3dgs} representation into adaptive voxel, and constructs an artifact-free static feature map by leveraging uncertainty prediction to filter out transient objects. The dashed region on the right presents results from baseline methods. In contrast, our approach achieves high-fidelity static scene reconstructions in the presence of dynamic elements.
    }
		\label{fig:face}
\end{figure}
%核心1.要表现出跟踪速度和建图速度不匹配导致uncertainty难以直接引入
\begin{abstract}
%3D Gaussian Splatting SLAM不断涌现，并持续突破以往难以实现的高质量场景构建与高实时渲染的技术瓶颈。然而，这些方法难以抵挡由动态物体引入的不一致性观测导致定位失效构建地图出现鬼影。为了解决这个挑战，我们提出了UP-SLAM，一个新颖的实时基于3d高斯的RGBD SLAM。具体地，我们设计了一种自适应结构化的3D高斯表征方法，通过引入概率八叉树结构能够自动管理高斯初始化和增删，避免设置繁琐复杂地阈值，提高渲染质量。其次，为了提高在动态环境下的定位精度和地图质量，我们提出了一种基于多特征的不确定度预测方法，通过融合多特征残差和浅层多层感知机实现实时鲁棒跟踪和高质量地图构建。我们验证UP-SLAM不仅能够实时定位和构建无鬼影地图，而且优于SOTA动态rgbd slam。
Recent 3D Gaussian Splatting (3DGS) techniques for Visual Simultaneous Localization and Mapping (SLAM) have significantly progressed in tracking and high-fidelity mapping. However, their sequential optimization framework and sensitivity to dynamic objects limit real-time performance and robustness in real-world scenarios. We present UP-SLAM, a real-time RGB-D SLAM system for dynamic environments that decouples tracking and mapping through a parallelized framework. A probabilistic octree is employed to manage Gaussian primitives adaptively, enabling efficient initialization and pruning without hand-crafted thresholds. To robustly filter dynamic regions during tracking, we propose a training-free uncertainty estimator that fuses multi-modal residuals to estimate per-pixel motion uncertainty, achieving open-set dynamic object handling without reliance on semantic labels. Furthermore, a temporal encoder is designed to enhance rendering quality. Concurrently, low-dimensional features are efficiently transformed via a shallow multilayer perceptron to construct DINO features, which are then employed to enrich the Gaussian field and improve the robustness of uncertainty prediction. Extensive experiments on multiple challenging datasets suggest that UP-SLAM outperforms state-of-the-art methods in both localization accuracy (by 59.8\%) and rendering quality (by 4.57 dB PSNR), while maintaining real-time performance and producing reusable, artifact-free static maps in dynamic environments. The project: \url{https://aczheng-cai.github.io/up_slam.github.io/}

\end{abstract}

\section{Introduction}

Visual Simultaneous Localization and Mapping (SLAM) is a core technology for embodied intelligence and virtual reality. Traditional SLAM algorithms typically assume static environments, which has facilitated the development of numerous effective systems~\cite{orbslam3,dso,plslam}. However, this assumption restricts the applicability of SLAM in dynamic real-world environments, thereby impeding advancements in robotics and related fields. Recent SLAM approaches~\cite{dynaslam,rldslam,dsslam} leverage object detection and multiple-view geometry theory to reduce the impact of dynamic objects. While these approaches enhance system robustness in dynamic environments, they heavily depend on prior knowledge of dynamic objects and the reliability of detection algorithms. 

Advances in high-fidelity scene representations, such as Neural Radiance Fields~\cite{nerf} (NeRF) and 3D Gaussian Splatting~\cite{3dgs} (3DGS), have motivated interest in introducing uncertainty modeling into 3D reconstruction. Recent studies~\cite{nerfonthego,wildgaussians,t3dgs} show that incorporating uncertainty prediction can significantly enhance robustness to transient scene elements. These uncertainty-aware models can achieve high-quality reconstructions even under intermittent occlusions. However, these methods depend on advantageous conditions, such as accurate camera poses and sparse viewpoints, which are challenging to achieve in SLAM systems using continuous frame inputs. 
%Additionally, the inconsistency between the frequencies of tracking and mapping complicates the direct incorporation of these methods into SLAM pipelines.

To address these challenges, a real-time RGB-D SLAM system named UP-SLAM is presented for robust pose estimation and static scene reconstruction in dynamic environments. Our approach compresses 3DGS into structured anchors encoded by multiple shallow multilayer perceptrons (MLPs). A probabilistic octree is introduced to enable adaptive adjustment of anchors to delete redundant anchors caused by dynamic objects. Furthermore, by decoupling motion mask generation from map optimization, UP-SLAM enables parallel tracking and mapping, supporting real-time localization. In the tracking process, we propose a training-free, optimization-based multi-modal consistency estimation method that fuses geometric cues with DINO features for effective dynamic object recognition. In the mapping process, to further enhance reconstruction under dynamic conditions, a temporal encoder that leverages sinusoidal positional encoding is designed to embed inter-frame information into the MLP, thereby increasing the representational capacity. In addition, the inconsistent appearance and motion of dynamic objects across frames provide valuable cues for uncertainty prediction. Therefore, robust DINO features are fed into a shallow MLP for per-pixel uncertainty estimation, enabling continuous motion mask refinement and enhancing reconstruction robustness.

Our primary contributions are as follows: (\textbf{i}) An uncertainty-aware parallel tracking and mapping framework is proposed to effectively mitigate dynamic disturbances without relying on predefined semantic annotations, thereby enabling the construction of high-quality, artifact-free static maps. (\textbf{ii}) We propose an adaptive structured 3DGS scene representation with a probabilistic octree, which supports automatic Gaussian primitive allocation or pruning in dynamic environments. This approach enhances localization accuracy and reduces model size. (\textbf{iii}) We integrate our approach into ORB-SLAM3~\cite{orbslam3} and perform comprehensive evaluations on multiple datasets. Additionally, we introduce a protocol for assessing rendering quality in dynamic environments, and we will release our datasets to the public.

%经过三十年的发展，视觉SLAM得到蓬勃发展出现了【】【】【】优秀算法，然而这些方法均忽略观测中速度分量即静态环境。动态物体的出现导致速度分量难以估计使得这些算法里程计出现漂移。为了解决这一问题，YOLO-SLAM【】【】【】利用更加轻量化的darknet-19替换了原来的darknet-53的yolo目标检测提高实时性但是检测框会框选出多余的静态区域导致定位精度下降。为此，DynaSLAM【】【】【】【】利用Mask-RCNN【】预先设定动态语义标签结合几何约束实现对异常观测的剔除。然而这些方法都需要提前设定标签，并且限制了适用性。为了摆脱设定标签的约束，PC-SLAM【】【】【】利用三维点之间空间相关性对动态特征点剔除，【】【】利用点云捕捉运动物体引起的不一致性，由于几何方法的鲁棒性不如深度学习方法导致难以应对复杂多变的环境。这些方法虽然实时性高但是鲁棒性不足适用范围受限，此外这些方法难以构建具有丰富信息的地图，为机器人导航等高级任务提供服务。然而我们将高维视觉特征dino高效的蒸馏到3dgs中实现特征地图构建为后续高级任务提供基础
%imap首次将以体渲染的nerf场景表征引入SLAM中，通过单个mlp和特征网格编码几何信息，实现隐式场景表征。但由于单个mlp的遗忘问题导致长期定位受到限制。随后NICESLAM【】针对长期定位的问题提出基于网格多层特征，实现更大场景的建图并且极大提高了实时性。VoxFusion【】，co-slam【】，eslam【】结合sdf进行混合表示，提高场景重建质量和计算效率。这些方法在静态环境下的效果令人印象深刻，但在动态环境下却发生了退化。最近，RodynSLAM采用与传统动态SLAM类似的语义分割先验和光流技术实现动态区域的过滤。forget【】提前训练一个预训练动态物体类别的分类器在跟踪过程中不断将根据地图和真值之间的残差识别出的物体特征喂给分类器使他能够学习到新的动态物体。然而，此类方法在很大程度上依赖于对物体类别的先验知识，导致在包含未知目标的真实环境中通用性较差，难以适应开放世界下的动态场景。
%nerf能够提供高保真重建但是渲染实时性较差，最近，基于gpu tile加速框架的新颖的3dg技术实现了极高频率的渲染速率。紧接着，出现了许多基于3dgs的方法【】【】【】。例如，splaTAM【】提出了基于轮廓mask的跟踪方法，并且基于几何对欠重建区域进行初始化高斯。类似同样初始化高斯的gsslam【】首次推导出了位姿梯度的解析解，并且实现粗到精的跟踪方法。但这些方法跟踪和建图耦合程度高导致实时性低难以应用于机器人。为此，phtotslam【】通过解耦跟踪和建图并且利用高实时性的orbslam3作为前端为训练图像提供位姿，为了实现高质量建图提出了图像金字塔优化方法，最终完成在嵌入式设备上实时跟踪和高质量建图。这些方法同样在动态环境下面临着挑战。为此，DG-SLAM【】结合分割和历史帧生成运动mask，实现摆脱依赖物体的先验信息。此外，gassidy【】利用分割和高斯混合模型对动态物体识别实现跟踪和高质量地图构建。这些方法前端不仅提供训练帧的位姿还提供动态物体的mask，但是由于mask过度依赖前端不灵活导致系统建图鲁棒性降低。wildgsslam是一个非常先进的单目动态slam系统，引入基于不确定度的动态物体识别方法实现了无鬼影高质量地图构建。这些方法跟踪和建图耦合程度高导致跟踪实时性大幅度降低。相对的，我们解耦跟踪和建图以提高跟踪实时性，并基于2D基础视觉模型dinov2特征实现mask的动态学习，实现了在动态环境下实时前端跟踪和无鬼影的高质量特征地图构建
\section{Related work}
\subsection{Traditional Visual SLAM}
Over the past few decades, visual SLAM has made remarkable progress, leading to the development of numerous outstanding algorithms~\cite{dso,orbslam3,plslam}. However, these methods generally assume a static environment and neglect the velocity components in observations, making them susceptible to severe drift when dynamic objects are present. To address this issue, dynamic SLAM methods~\cite{yoloslam,dsslam} based on object detection or semantic segmentation have been proposed.  DynaSLAM~\cite{dynaslam} leverages predefined semantic labels from Mask R-CNN~\cite{maskrcnn}, combined with geometric constraints, to reduce the influence of dynamic objects. However, this dependence on prior labels significantly limits the generalizability of the method. To reduce reliance on semantic priors, the method~\cite{pcslam} takes advantage of spatial correlations between 3D points to eliminate dynamic features, while other methods~\cite{spwslam,cprslam} detect inconsistencies between point clouds to localize moving objects. Most existing methods emphasize tracking robustness but neglect the construction of semantically enriched and reusable maps, which are critical for high-level tasks such as navigation and scene understanding. To bridge this gap, we design to efficiently distill high-dimensional visual features from DINOv2~\cite{dinov2} into the 3DGS representation, enabling the construction of a dense, feature-rich map that serves as a reliable foundation for downstream robotic tasks.

\subsection{NeRF and 3DGS SLAM}
\paragraph{NeRF SLAM.}
Neural implicit SLAM methods have recently attracted increasing attention for their capability to reconstruct high-fidelity scenes with continuous representations. iMAP\cite{imap} is the first to introduce volumetric NeRF scene representations into SLAM, encoding geometry through a single MLP and feature grid for implicit mapping. However, the long-term localization performance is hindered by the catastrophic forgetting issue associated with single-network representations. To address this, NICE-SLAM~\cite{niceslam} proposes a hierarchical grid-based feature structure, significantly improving scalability and real-time performance for large-scale scene reconstruction. Subsequent works such as Vox-Fusion\cite{voxfusion}, Co-SLAM~\cite{coslam}, and ESLAM~\cite{eslam} incorporate signed distance fields into hybrid representations, achieving notable gains in reconstruction quality and computational efficiency. While these methods demonstrate impressive performance in static scenes, they tend to degrade under dynamic conditions. More recent approaches have sought to extend NeRF-based SLAM to dynamic environments. For example, RodynSLAM~\cite{rodynslam} combines semantic priors and an optical flow estimator to mask dynamic regions. The method~\cite{forget} pre-trains a dynamic object classifier and incrementally updates it by feeding features from objects identified through residuals between the map and ground truth. This enables the system to learn new dynamic features over time. However, such approaches heavily rely on prior knowledge of object classes, making them less generalizable to real-world open-set environments that contain unknown dynamic objects.
\paragraph{3DGS SLAM.}
Recently, the emergence of 3DGS techniques, leveraging GPU tile-based acceleration frameworks, has enabled extremely high-frequency rendering. Several 3DGS-based SLAM systems have been proposed~\cite{cgslam,gsorb,gs3slam,gaussianslam,gsslam}. For instance, SplaTAM~\cite{splatam} proposes a silhouette tracking strategy and uses geometric cues to initialize Gaussians in under-reconstructed regions. However, both methods exhibit a high degree of coupling between tracking and mapping, which limits their real-time performance and hampers deployment in robotic platforms. To address this, Photo-SLAM~\cite{photoslam} decouples tracking and mapping by employing the real-time ORB-SLAM3 tracking to provide camera poses for training images. It further proposes an image pyramid optimization strategy to improve reconstruction quality, ultimately achieving real-time tracking and high-quality mapping on embedded devices. Nevertheless, these methods still face challenges in dynamic environments. DG-SLAM~\cite{dgslam} addresses this by combining semantic segmentation with motion masks derived from spatial geometry consistency across frames, thereby removing the need for prior knowledge of object classes. Gassidy~\cite{gassidy} employs segmentation and a Gaussian mixture model to mask dynamic objects for robust tracking and high-quality map reconstruction. These methods obtain a fixed motion mask from the tracking, which can reduce the robustness of mapping. WildGS-SLAM~\cite{wildslam} represents a recent advancement in monocular dynamic SLAM, introducing an uncertainty-aware dynamic object recognition strategy to build high-quality, artifact-free maps. However, the tight coupling between tracking and mapping in these methods compromises real-time performance. In contrast, tracking and mapping are decoupled in our method to improve efficiency, and DINO features are fed into a shallow MLP to support continuous refinement of the motion mask, thereby supporting real-time tracking and artifact-free, high-quality map reconstruction in challenging dynamic environments.

\section{Approach}
\begin{figure}[t]
    \centering
    \includegraphics[scale=0.45]{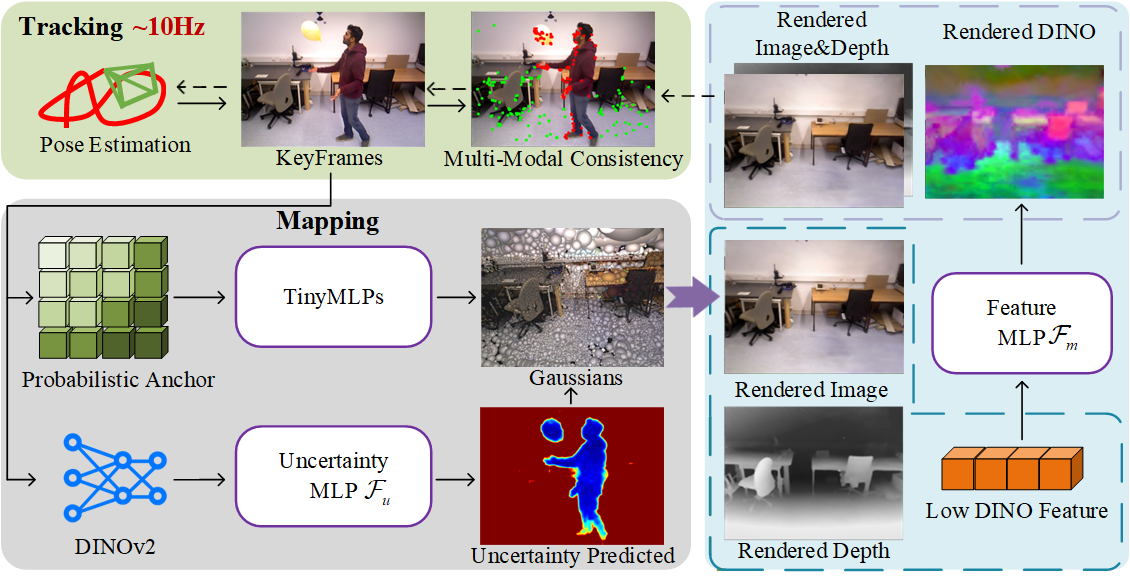}
    \caption{\textbf{System overview.} UP-SLAM is a parallel tracking and mapping system that enables real-time localization and high-fidelity, artifact-free map reconstruction.
    }
		\label{fig:pipeline}
        \vspace{-4mm}
\end{figure}

An overview of the UP-SLAM system is shown in Fig.~\ref{fig:pipeline}. UP-SLAM takes a sequence of RGB and depth images $\{\mathcal{D},\mathcal{I}\} \in \mathbb{R}^{H \times W}$ as input and adopts a parallelized tracking and mapping architecture to enhance overall efficiency. In the tracking thread (Sec.~\ref{sec:tracking}), the system performs real-time localization and generates keyframes for mapping. Dynamic region detection is guided by multi-modal residuals propagated from the mapping thread, enabling robust and real-time tracking. The mapping thread (Sec.~\ref{sec:mapping}) employs probabilistic anchors to construct an adaptively structured 3DGS representation, which reduces model size while improving reconstruction quality. To improve mapping quality in dynamic environments, robust 2D visual features extracted from DINOv2 are distilled into the 3DGS representation to construct multi-modal residuals, which supervise a shallow MLP for per-pixel uncertainty prediction and enable continuous refinement of the motion mask. 

%输入rgb和深度图序列通过跟踪生成关键帧送入到后端建图中，为前端多模态一致性估计提供残差，然后通过mask转化器生成mask剔除动态特征点。后端通过增量概率八叉树管理anchor，通过多个MLP解码出gs的属性包括颜色，透明度，旋转，尺度，低维度特征。通过渲染得到色彩图和深度图以及低维度特征，然后通过mlp将低维度特征转化至高维度dino特征。随着后端地图不断地优化通过dinov2特征为输入的mlp预测的不确定度将会越来越准确，不断调整mask优化高斯参数最终构建静态无鬼影的特征地图。
%在动态环境下捕获一系列图像和深度图。UP-SLAM解耦前后端，前端跟踪只为后端建图提供关键帧，后端只为前端提供残差。通过残差引导实现多模态一致性估计出动态区域，实现在动态环境下的准确鲁棒实时的跟踪。引入概率八叉树数据结构实现自适应结构化3dgs建图，进一步压缩模型大小提高建图质量。为了增强残差的辨识度，将2d视觉特征蒸馏到3dgs中。基于DINOv2特征输入、以多模态残差为监督的不确定度mlp预测有较好的鲁棒性，并且通过不断预测调整mask优化高斯参数最终构建静态无鬼影的特征地图。

\subsection{Preliminaries}

Following the vanilla 3DGS~\cite{3dgs} framework, the entire scene is represented by a set of anisotropic Gaussian ellipsoids $\textbf{G}$:
\begin{equation}
{\textbf{G}} = \{ {G_i}:(\mu_{i} ,o_{i},c_{i},\Sigma_{i} )|i = 1,...,N\},
\end{equation}
where each Gaussian is defined by its color $c \in \mathbb{R}^{3}$, opacity $o \in [0,1]$, position $\mu \in \mathbb{R}^{3}$, and covariance matrix $\Sigma \in \mathbb{R}^{3 \times 3}$. The covariance matrix $\Sigma$ is decomposed as $\Sigma_i  = RS{S^T}{R^T}$, where $S$ is a scale matrix and $R$ is a rotation matrix, to ensure positive semi-definiteness.

The camera-to-world transformation $T_{wc}$ is obtained via pose estimation method, after which each 3D Gaussian point $G_{i}$ is projected onto the image plane for rendering, as follows:
\begin{equation}
\Sigma ' = JT_{wc}^{ - 1}\Sigma T_{wc}^{ - T}{J^T},
\end{equation}
where $J$ denotes the Jacobian matrix of the affine approximation to the projective function. Following the $\alpha$-blending technique, the rendered color $\tilde C$ and depth $\tilde D$ of each pixel are computed by accumulating the contributions of Gaussians along the ray. In addition, the accumulated transmittance $\tilde T$ is rendered to determine visibility, as formulated below:
\begin{equation}
\{ \tilde C,\tilde D\}  = \sum\limits_{i = 1}^N {\{ {c_i},{z_i}\} {\sigma _i}} \prod\limits_{j = 1}^{i - 1} {(1 - {\sigma _j})} ,\tilde T = \sum\limits_{i = 1}^N {{\sigma _i}} \prod\limits_{j = 1}^{i - 1} {(1 - {\sigma _j})},
\end{equation}
where $c_i$ represents the color of the i-th 3D Gaussian, the density $\sigma_{i}$ is determined by both the Gaussian distribution function and the learned opacity $o_{i}$, the $z_{i}$ denotes the Gaussian depth value in the camera coordinate.
In the optimization of Gaussian parameters, we incorporate geometric supervision as follows:
\begin{equation}
\label{equ:lgloss}
{\mathcal{L}_{g}} =  \lambda _1(\lambda \left\| {\tilde C - C} \right\|_2^2 + (1 - \lambda)(1 - \textbf{SSIM}(\tilde C,C))) + \lambda _2\left\| {\tilde D - D} \right\|_2^2,
\end{equation}
where $\textbf{SSIM}$ is structural similarity index measure~\cite{ssim}, $\{\lambda\}$ are hyperparameters. $C$ and $D$ denote the ground-truth color and depth, respectively.

\subsection{Uncertainty Model}
\label{sec:unc}
%以往的动态slam工作识别动态干扰均是前后端分离进行的未能将地图信息有效利用起来，进而存在许多限制，如提设定语义标签【】，训练分类器【】等。我们解决这个问题受【】【】的启发，通过基于贝叶斯学习框架，令模型输出每个像素不确定程度的高斯分布而不是单个值，他无需预训练。对于每个像素，计算渲染和真值之间的残差R，xita是不确定度。每个像素的损失是正态分布的负对数似然：

The effectiveness of uncertainty prediction in filtering transient objects has been well demonstrated in in-the-wild 3D reconstruction tasks~\cite{nerfonthego,wildgaussians,robustnerf}. The uncertainty is modeled using a Bayesian learning framework, where the model predicts a Gaussian distribution to represent the uncertainty of each pixel, rather than outputting a single deterministic value. For each pixel, we compute the residual $R$ between the rendered value and the ground truth, with $\sigma$ denoting the predicted uncertainty. The loss for each pixel is defined as the negative log-likelihood of a normal distribution:
\begin{equation}
{{\cal L}_u} =  - \log (\frac{1}{{\sqrt {2\pi {\sigma ^2}} }}\exp ( - \frac{R}{{2{\sigma ^2}}})) = \frac{R}{{2{\sigma ^2}}} + {\lambda _3}\log \sigma  .
\end{equation}
The first term is regularized by the second term, which corresponds to the log-partition function of the normal distribution and prevents a trivial minimum at $\sigma = \infty$~\cite{nerfw}.

The DINO features are inherently robust to appearance variations across frames~\cite{wildgaussians}, making them well-suited for dynamic scenes with inconsistent appearance features. Therefore, DINO features are incorporated into both color and depth information to achieve joint constraints across appearance, geometry, and semantics. Additionally, we employ the accumulated transmittance $\tilde{T}$ as a visibility mask to prevent low-opacity regions from contributing. The total residuals $R$ are defined as:
\begin{equation}
\label{equ:residual}
R = (\tilde T < 0.1)(\lambda _1^t\left| {\tilde C - C} \right| + \lambda _2^t\left| {\tilde D - (D \otimes {\bf{B}})} \right| + \lambda _3^t{\left\lceil {1 - \frac{{F \cdot \hat F}}{{{{\left\| F \right\|}_2}{{\left\| {\hat F} \right\|}_2}}}} \right\rceil ^1}).
\end{equation}
The $\textbf{B}$ is a 3×3 box filter applied to depth via convolution ($\otimes$), and $\left\lceil {} \right\rceil^1 $ indicates that the output is capped at 1. $F$ denotes the visual features extracted by DINOv2~\cite{dinov2}, while $\hat F$ signifies the rendered high-dimensional visual features. Since DINOv2 is defined per image patch, we perform bilinear interpolation to upsample it to the image size for similarity calculation.

\subsection{Mapping}
\label{sec:mapping}
%%%%%%%%%%%%%%%%%%%%%%%%%%%%%%%%%%%%%%%%%%%%%%%%%%%%%%%%%%%%%%%%%%
%以视频为序列的slam技术需要快速识别欠重建区域并且初始化新的高斯点以提高跟踪效率。目前，许多3DGS SLAM需要设置繁琐的阈值识别欠重建区域【】【】【】，在动态环境下这些初始化方法将会受到挑战。采用不合理的阈值会增加显存负担降低计算效率甚至会导致渲染质量下降。受【oct】【sca】的启发，我们设计了一种基于增量式概率栅格的3DGS，避免了设置繁琐的阈值和手动管理高斯点的增删。将k个高斯属性从anchor特征，相机光心和anchor之间的相对方向和距离通过mlp解码。[sca]对色彩添加外貌嵌入来提升渲染质量，【scslma】同样的仅对色彩添加了以位姿为输入的mlp提升渲染质量，不幸地，旋转矩阵属于具有约束的流形so3导致mlp学习非常困难。针对于SLAM具有时间序列的图像且位姿与时间高度相关，我们发展额外的时间编码进一步提高渲染质量。因此，我们将图像序列输入到周期性编码中进行线性化：

%最终描述为。。。。，anchor解码出的颜色可表示为：

%类似的F,F,F。此外，我们的anchor拥有概率属性，概率值表示anchor的动态程度这将更加符合动态环境,栅格概率更新公式如下：
%更多详细的请参考【】。

%%%%%%%%%%%%%%%%%%%%%%%%%%%%%%%%%%%%%%%%%%%%%%%%%%%%%%%%%%%%%%%%%%

%%%%%%%%%%%%%%%%%%%%%%%%%%%%%%%%%%%%%%%%%%%%%%%%%%%%%%%%%%%%%%%%%%
%为了提高在连续帧中对动态物体的区分程度，我们采用基于可微3DGS框架的渲染管道类似于颜色和深度来渲染视觉特征，但是高维度渲染会大幅度降低计算效率。受【gs3】【feat】的启发，为了防止过拟合和提高计算效率，每个anchor解码出k个拥有Nl维低维视觉特征的高斯，
%然后，通过浅层mlp恢复到Nh维高维视觉特征，
%由于Nl<<Nh，在不降低优化效率的同时将视觉特征嵌入到3DGS中
%%%%%%%%%%%%%%%%%%%%%%%%%%%%%%%%%%%%%%%%%%%%%%%%%%%%%%%%%%%%%%%%%%
%由于建图中与跟踪不同，不确定度需要不断地优化构建mask并且实时性要求低于跟踪但对动态物体识别的要求更高。因此，我们仅仅将跟踪中使用的全分辨率不确定度用一个以dino特征为输入的mlp预测不确定度来代替。将优化后的不确定度转化成二进制mask

\paragraph{Adaptively Structured Gaussian.}
3DGS SLAM techniques require the rapid identification of under-reconstructed regions and the initialization of new Gaussian primitives to improve tracking efficiency. Currently, many 3DGS SLAM systems rely on manually tuned thresholds to detect under-reconstructed regions\cite{gsorb,splatam,gsslam}. However, The threshold-dependent initialization strategies become increasingly unreliable in dynamic environments. Inappropriate threshold settings may lead to excessive GPU memory consumption, reduced computational efficiency, and even degraded rendering quality.

An incremental probabilistic anchor method is proposed to achieve adaptively structured 3DGS, eliminating the need for complex threshold tuning and manual management of Gaussian primitives. Specifically, we decode the $k$ Gaussian attributes from the anchor features ${\hat f}_v$, as well as the relative direction $\delta _{vc}$ and distance ${\vec d}_{vc}$ between the camera center and the anchor, using an MLP~\cite{scaffold}. In contrast, our anchors are equipped with probabilistic attributes, where the probability value reflects the degree of motion at each anchor. This probabilistic representation is more suitable for dynamic environments. The probabilistic anchor update equation is as follows~\cite{octomap}:
\begin{equation}
P(n|{z_{1:t}}) = {[1 + \frac{{1 - P(n|{z_t})}}{{P(n|{z_t})}}\frac{{1 - P(n|{z_{1:t - 1}})}}{{P(n|{z_{1:t - 1}})}}\frac{{P(n)}}{{P(n)}}]^{ - 1}}.
\end{equation}
This update equation is based on Bayes theorem and requires a prior probability $P(n)$, the current observation $z_{t}$, and the likelihood model $P(n|z_{1:t-1})$ to update the dynamic probability of each anchor. $P(n \mid z_t)$ denotes the probability that anchor $n$ is occupied, given the observation $z_t$.

\paragraph{Temporal Encoding.}
Methods such as~\cite{scaffold,wildgs} introduce appearance embeddings into the color prediction to improve rendering quality in wild reconstruction. These methods are primarily designed to improve rendering quality through better representation learning. In the context of SLAM, the method~\cite{scaffoldslam} leverages the pose as an additional input to the MLP, utilizing the characteristics of SLAM to enhance performance.  
However, since the rotation matrix lies on the special orthogonal group $SO(3)$, a non-Euclidean manifold with nonlinear constraints, conventional MLPs struggle to model rotational variations effectively~\cite{rotation}, leading to suboptimal performance. Given that SLAM operates on temporally correlated image sequences where pose evolution is time-dependent, we propose a temporal encoding method to further enhance rendering quality. Specifically, each sequence $t$ is mapped to a temporal embedding ${\ell _t} = \{ \sin (\pi t ),\cos ( \pi t ) \} \in {\mathbb{R}^2} $, which improves the representational capacity of all MLPs. For example, the color \{c\} is predicted using an MLP conditioned on both spatial and temporal features:
\begin{equation}
\{ {c_0},...,{c_{k - 1}}\}  = {\mathcal{F}_c}({{\hat f}_v},{\delta _{vc}},{{\vec d}_{vc}},{\ell _t}).
\end{equation}
Similarly, opacity $\{o\}$, rotation $\{q\}$, and scale $\{s\}$ are each predicted by their individual MLPs.

\paragraph{Visual Feature.}
The inclusion of high-dimensional visual features significantly expands the Gaussian optimization space, leading to higher memory consumption and reduced computational efficiency. Inspired by~\cite{featuregs,gs3slam}, anchor features are employed to decode low-dimensional Gaussian visual attributes $\{f\} \in \mathbb{R}^{k \times N_{l}}$ via an MLP $\mathcal{F}_d$:
\begin{equation}
\label{equ:dinodecoder}
\{ {f_0},...,{f_{k - 1}}\}  = {\mathcal{F}_d}({{\hat f}_v},{\delta _{vc}},{{\vec d}_{vc}},{\ell _t}),
\end{equation}
Similar to color rendering, the low-dimensional features are rendered through the 3DGS framework, yielding the rendered feature representation $\tilde F$, as defined below:
\begin{equation}
\tilde F = \sum\limits_{i = 1}^N {{f_i}{\sigma _i}}\prod\limits_{j = 1}^{i - 1} {(1 - {\sigma _j})}.
\end{equation}
To align the low-dimensional Gaussian parameters with the high-dimensional $N_{h}$ visual features, we employ a shallow MLP $\mathcal{F}_m$ to map them into a higher-dimensional space and obtain high-dimensional visual features $\hat F$:
\begin{equation}
\hat F = {\mathcal{F}_m}({\tilde F}) \in \mathbb{R}^{N_{h}}.
\end{equation}
We supervise the learning of DINO features $F$ and rendering features $\hat{F}$ through a loss function $\mathcal{L}_d$:
\begin{equation}
\mathcal{L}_d = \frac{1}{N_d}\sum\limits_{i = 0}^{N_d} \left(1 - \frac{F_i\cdot\hat F_i}{\left\| F_i \right\|_2 \left\| \hat{F}_i \right\|_2}\right), 
\end{equation}
where $N_d$ is the feature dimension of DINO, $i$ is the i-th vector. Since $N_{l} << N_{h}$, visual features $\hat F$ are efficiently distilled into the 3DGS representation, preserving optimization efficiency while reducing memory and computational overhead.

\paragraph{Uncertainty Prediction for Mapping.} 
In traditional SLAM systems, the tracking module performs real-time localization, while the mapping module focuses on mitigating accumulated drift and optimizing the map. This paradigm is followed in our work, where the mapping process is extended to include not only the optimization of the static scene representation but also the refinement of the motion mask, thereby enhancing both robustness and reconstruction quality. Specifically, DINO features are fed into an MLP $\mathcal{F}_u$, which predicts per-pixel uncertainty:
 \begin{equation}
\sigma = \mathcal{F}_u(F),
\end{equation}
and the parameters of the MLP $\mathcal{F}_u$ are optimized under the supervision of the loss function $L_u$. The uncertainty map is then binarized to generate a motion mask $M = \delta (2{\sigma ^2} > 1)$, where $\delta$ denotes an indicator function that returns true when the condition is satisfied. To ensure multimodal consistency of $\textbf{G}$ in dynamic environments, $\mathcal{L}$ encompasses constraints related to appearance, semantics, geometric and motion mask as follows:
\begin{equation}
\mathcal{L} = {M}(\mathcal{L}_{g}+\lambda _4 \mathcal{L}_d) + \lambda _5\bar s,
\end{equation}
where $\bar s$ denotes the mean scale, introduced to prevent scale explosion. During each Gaussian optimization iteration, the uncertainty MLP $\mathcal{F}_u$ is optimized simultaneously. However, the gradient flows from the mapping loss $\mathcal{L}$ and the uncertainty loss $\mathcal{L}_u$ are kept separate to ensure independent parameter updates.

\subsection{Tracking}
\label{sec:tracking}
%我们的前端跟踪基于特征点的orbslam3，与【】【】【】不同，为了提高实时性我们不仅只处理关键帧而且在跟踪中不会利用mlp估计不确定度，因为这是一次性操作。为了完成这一目标，我们需要设置一个能够稳定鲁棒训练的残差。【】利用DSSIM计算损失表现出了稳定训练动态的效果，【】证明了利用dino特征对外貌变化具有更好地鲁棒性，因此为了减少动态物体的影响，我们设计联合多模态残差损失。具体地，对输入的关键帧进行dino特征的提取，并且初始化全分辨率不确定度xita，将颜色、深度与DINO特征融合于统一优化目标中，从而兼顾外观、几何与语义信息的协同约束，其总残差R：
%其中深度会与3x3的盒式滤波器B进行卷积（*），S是相似度计算函数，在相似度计算之后采用双向插值进行上采样。利用累计折射率作为mask防止因欠重建区域对残差的干扰。
%随后将优化后的不确定分数转化为二进制mask,并且与yolo分割计算Iou获取分割区域，这个mask会被用于剔除特征点防止转化成landmark实现鲁棒位姿估计T*：
%注意虽然yolo分割是close-set的但并不影响upslam泛化到open-set的动态区域识别的能力如图。引入分割是因为slam中对于每一帧都需要及时做出动态区域的剔除，一旦受到非正常观测就会导致定位失效，对于具身智能来说这将是灾难性的。然而由于初始状态时动态物体在连续帧中存在极小的运动变化难以识别出特别是非刚性对象，因此yolo分割只是为了做补全。
Previous methods such as~\cite{wildslam,forget,dgslam} follow a sequential tracking-mapping pipeline, in which camera pose estimation is followed by scene representation optimization, and dynamic object recognition is typically performed after map convergence. However, this tight coupling between modules limits the ability to achieve real-time localization in dynamic environments.

Therefore, to achieve a parallel tracking and mapping framework, a training-free estimator is proposed to decouple motion mask generation from global scene optimization. By exploiting the fast rendering capabilities of 3DGS, multi-modal residuals $R$ are computed and fed in real time into the following objective function $\xi$, from which an uncertainty map $\sigma$ is optimized:
\begin{equation}
\xi(\sigma)  = \mathop {\min }  \frac{1}{{HW}}\sum\limits_{i = 1}^H {\sum\limits_{j = 1}^W {\frac{1}{2}\left( {\frac{{{R_{ij}}}}{{{\sigma ^2}}} + \log \sigma } \right)} }.
\end{equation}
The uncertainty map is subsequently thresholded to generate a motion mask, which is used to filter out dynamic keypoints from the keyframes, preventing them from being converted into landmarks.

\begin{figure}[t]
 \vspace{-6mm}
    \centering
    \includegraphics[scale=0.47]{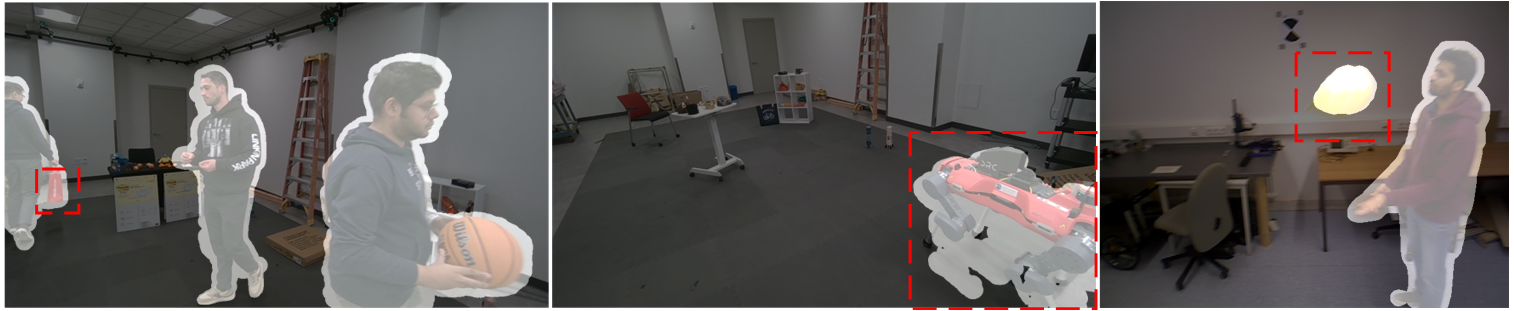}
     \vspace{-2mm}
    \caption{\textbf{Open-set capability in tracking.} The categories within the red box are all without pre-training, and the bright regions indicate the detected dynamic objects.}
		\label{fig:openset}
        \vspace{-4mm}
\end{figure}

During initialization, the complete extraction of dynamic regions is essential for feature-based SLAM, as even partial inclusion of dynamic features can lead to long-term drift or tracking failure. To enhance mask completeness, we refine motion mask by computing its intersection over union with segmentation results from YOLOv8-seg~\cite{yolov8}, ensuring more thorough exclusion of dynamic regions. While YOLOv8-seg is trained on a closed set of categories, our residual-guided refinement strategy allows UP-SLAM to generalize effectively to untrained dynamic objects mask, as illustrated in Fig.~\ref{fig:openset}.

% The segmentation module is employed to ensure the timely removal of dynamic regions, which is crucial for feature-based SLAM. The introduction of potential dynamic observations can lead to tracking failure, which is catastrophic for robotic applications.  This refined mask is then used to filter out feature points, preventing them from being converted into landmarks and thus enabling robust pose estimation $T^*$:
% \begin{equation}
% {T^*} = \mathop {\min }\limits_{\{R,t\} \in T_{wc}} (\sum\limits_{i \in 
%  \mathcal{P}} {\omega (\left\| {{p_i} - {M_t}(\prod (T_{wc},{P_i}))} \right\|_2^2)} ),
% \end{equation}

% Here, $P_{i}$ and $p_{i}$ represent the corresponding landmark and observation, respectively, while $\omega$ denotes the robust kernel function. The $\mathcal{P}$ is landmarks and $\prod$ is project function. Although YOLOv8~\cite{yolov8} segmentation is closed-set, it does not hinder the ability of UP-SLAM to generalize to open-set dynamic region detection, as illustrated in Fig.~\ref{fig:openset}. 

% However, due to the minimal motion variation of dynamic objects in consecutive frames at the initial state, it is difficult to identify them, especially non-rigid objects. Therefore, YOLO segmentation serves as a complementary step to handle such challenging cases.

%在表123中，我们的方法在baselin中表现突出，由于在表2中只有人是动态物体所以dynaslam的设定动态目标的方式起到了很好的效果，但是在表1中可以看出有些退化在表3中更是如此。相比于同样基于3dgs的动态slam，dgslam和gassidy，他们也具有openset识别能力但是定位精度均不如我们的方法，我们的方法在定位精度上比与SOTA的dg-slam平均提高了59.8%。潜在的表示出我们算法强大的鲁棒性。
%由于动态mask的引入可能导致位姿估计精度发生退化，因此，为了证明我们方法的鲁棒性，我们在公共静态真实环境的数据集SCANNET进行评估。我们的方法和针对静态环境的设计的SLAM相比定位精度平均提升了10%。与此同时，我们与同样针对动态环境设计的DG-SLAM相比平均精度提升了8.1%。这证明我们方法无论针对动态还是静态环境都有较好的鲁棒性和准确的定位精度。
%为了体现不同组件的作用，我们进行消融实验如表1所示，时间编码主要贡献是提升渲染质量。分割主要影响定位和渲染质量，因为非刚性物体的存在，在前期upslam难以有效的分割出整个动态区域导致部分潜在动态特征点被转化成地图用于后续的跟踪。概率更新anchor组件极大的压缩了模型的大小使其更适合在嵌入式设备上复用，此外由于anchor不更新导致gs点不会被删除，地图更新缓慢难以有效快速的反馈准确的残差给前端,所以定位精度有所下降。时间编码和分割对dino特征蒸馏效果都有一定的影响，值得注意的是，upslam能够将相似度提高到接近80%，这扩展了下流任务的应用比如物体级导航和语义理解等。
%表1是时间分析实验，由于我们解耦了前后端每帧速度可达到12hz足以实现机器人在动态环境下的实时定位的要求,我们和Wil设置相同的迭代次数，总的时间比wild快2倍并且在refine可达到30hz的训练速度。此外我们对模型也做出了极致的压缩。

\section{Experiments}

\begin{table}[b]
 \vspace{-4mm}
    \renewcommand{\arraystretch}{1.0}
    \setlength{\tabcolsep}{1.2mm}
    \centering
    \caption{\textbf{Tracking results on Bonn RGB-D dataset.} (ATE RMSE$\downarrow$[cm])}
    \small
    \begin{tabular}{c|cccccc|c}
    \hline
        Method & Balloon & Balloon2 & Ball\_track & Ps\_track & Ps\_track2 & Mv\_box2 & Avg. \\ \hline
        ORB-SLAM3 & 5.8 & 17.7 & \underline{3.1} & 70.7 & 77.9 & \underline{3.5} & 29.78 \\ 
        DynaSLAM & \underline{3.0} & \underline{2.9} & 4.9 & 6.1 & 7.8 & 3.9 & \underline{4.76} \\ 
        ESLAM & 22.6 & 36.2 & 12.4 & 48 & 51.4 & 17.7 & 31.38 \\ 
        RoDyn-SLAM & 7.9 & 11.5 & 13.3 & 14.5 & 13.8 & 12.6 & 12.26 \\  
        Photo-SLAM & 6.9 & 26 & 3.2 & 76.4 & 87.4 & 3.6 & 33.91 \\
        GS-SLAM & 37.5 & 26.8 & 31.9 & 46.8 & 50.4 & 4.8 & 33.03 \\ 
        DG-SLAM & 3.7 & 4.1 & 10 & \underline{4.5} & \underline{6.9} & 3.5 & 5.45 \\ \hline
        UP-SLAM(Our) &\textbf{2.8} & \textbf{2.7} & \textbf{2.9} & \textbf{4.0} & \textbf{3.6} & \textbf{3.2} & \textbf{3.2} \\ \hline
    \end{tabular}
    
\label{tab:ate_bonn}
\vspace{-4mm}

\end{table}
\begin{table}[t]
    \renewcommand{\arraystretch}{1.2}
    \setlength{\tabcolsep}{0.3mm}
    \centering
    \caption{\textbf{Tracking results on MoCap RGB-D dataset.}  "X" denotes a tracking failure.  (ATE RMSE$\downarrow$[cm])}
    \small
    \begin{tabular}{c|cccccccccc|c}
    \hline
        Method & ANY1 & ANY2 & Ball & Crowd & Person & Racket & Stones & Table1 & Table2 & Umb. & Avg. \\ \hline
        DynaSLAM & 1.6 & \textbf{0.5} & \textbf{0.5} & 1.7 & \textbf{0.5} & \textbf{0.8} & 2.1 & \underline{1.2} & \underline{34.8} & 34.7 & 7.84 \\ 
        
        NICE-SLAM & X & 123.6 & 21.1 & X & 150.2 & X & 134.4 & 138.4 & X & 23.8 & - \\ 
        Photo-SLAM & 79.5 & 11.8 & 50.3 & 105.9 & 27.5 & 38.23 & 113.5 & 39.1 & 64.8 & 84 & 61.46 \\
         
        DG-SLAM & \underline{1.2} & 2.1 & 0.8 & \underline{1.3} & 1.5 & 1.6 & \underline{1.5} & 2 & 57.9 & \underline{1.35} & \underline{7.06} \\ \hline
        UP-SLAM(Our) & \textbf{0.4} & \underline{0.6} & \underline{0.6} & \textbf{1.1} & \underline{1.1} & \underline{0.9} & \textbf{1.0} & \textbf{0.7} & \textbf{3.6} & \textbf{0.8} & \textbf{1.08} \\ \hline
    \end{tabular}

\label{tab:ate_wild}
 \vspace{-2mm}
\end{table}
\begin{table}[t]
    \centering
    \caption{\textbf{Tracking results on TUM RGB-D dataset.} (ATE RMSE$\downarrow$ [cm])}
\small
\renewcommand{\arraystretch}{1.0}
\setlength{\tabcolsep}{1.0mm}
\begin{tabular}{c|ccccc|c}
\hline
Method     & Fr3/w/xyz & Fr3/w/half & Fr3/w/static & Fr3/s/xyz & Fr2/desk\_person & Avg.          \\ \hline
ORB-SLAM3  & 28.1      & 30.5       & 2.0          & 1.0       & 1.5              & 12.62         \\ 
DynaSLAM   & \textbf{1.5}  & 2.9        & 0.7          & 1.6       & \underline{0.9}              & 1.52    \\
Co-SLAM & 51.8 & 105.1 & 49.5 & 6 & 7.6 & 44 \\ 

RoDyn-SLAM & 8.3       & 5.6        & 1.7          & 5.1       & 5.6              & 5.26          \\ 
Photo-SLAM & 60.4      & 35.7       & 13.7         & \underline{1.0} & \textbf{0.6}              & 22.28         \\
DG-SLAM    & 1.7       & \textbf{1.8}  & \textbf{0.7}  & 1.0       & 3.2              & 1.68          \\\hline
UP-SLAM(Our) & \underline{1.6} & \underline{2.6}  & \textbf{0.7}    & \textbf{0.9}  & 1.3              & \textbf{1.42} \\\hline
\end{tabular}

\label{tab:ate_tum}
\vspace{-4mm}
\end{table}
 
\subsection{Experimental Setup}
\label{sec:exp}

% \paragraph{Dataset.}
% %我们评估我们的方法在TUM RGB-D Dataset，Bonn RGB-D Dynamic Dataset and Wild-SLAM MoCap Dataset.在TUM【】中主要动态物体是人，在bonn中除人以外存在少部分其他动态物体如气球、箱子等，这些数据集的图像分辨率均为480x640。在Wild-SLAM MoCap【】图像分辨率是1280x720并且存在除人以外大量非结构化的动态对象，如机器狗，石头等。这对需要提前设定动态对象的动态SLAM存在巨大的挑战。
% We evaluate our method on the TUM RGB-D Dataset~\cite{tum}, the Bonn RGB-D Dynamic Dataset~\cite{bonn}, and the WildGS-SLAM MoCap Dataset~\cite{wildslam}. 

% In the TUM dataset, the primary dynamic objects are humans. In the Bonn dataset, in addition to humans, a small number of other dynamic objects, such as balloons and boxes, are present. The image resolution for both datasets is 480×640.In contrast, the WildGS-SLAM MoCap dataset features a higher image resolution of 1280×720 and contains a large number of unstructured dynamic objects beyond humans, such as robotic dogs, rocks, and other irregular entities. This poses significant challenges for dynamic SLAM methods that rely on predefined dynamic object categories.

To demonstrate the competitiveness of our approach, we compare it against 16 methods, categorized as follows:
(a) Classic SLAM methods: ORB-SLAM3~\cite{orbslam3};
(b) Classic dynamic SLAM methods: ReFusion~\cite{refusion}, DynaSLAM~\cite{dynaslam}, EM-Fusion~\cite{emfusion};
(c) NeRF-based SLAM methods: iMAP~\cite{imap}, NICE-SLAM~\cite{niceslam}, Vox-Fusion~\cite{voxfusion}, Co-SLAM~\cite{coslam}, ESLAM~\cite{eslam};
(d) NeRF-based dynamic SLAM: RoDyn-SLAM~\cite{rodynslam};
(e) 3DGS-based SLAM: Photo-SLAM~\cite{photoslam}, GS-SLAM~\cite{gsslam}, SplaTAM~\cite{splatam};
(f) 3DGS-based dynamic SLAM methods: DG-SLAM~\cite{dgslam}, Gassidy~\cite{gassidy}, WildGS-SLAM~\cite{wildslam}.
All methods are evaluated using dynamic datasets, specifically the TUM RGB-D Dataset~\cite{tum}, the Bonn RGB-D Dataset~\cite{bonn}, and the MoCap RGB-D Dataset~\cite{wildslam}, in addition to a static environment dataset, the ScanNet Dataset~\cite{scannet}. We report original results for non-open-source methods, and average results over five runs for open-source ones. \textbf{Bold} is the best result, and \underline{underline} is the second best result. We select representative baselines from each category. Additional experimental results, dataset details, evaluation metrics, limitations and implementation details are provided in Appendix~\ref{sec:appendix}.
\begin{table}[b]
\vspace{-4mm}
    \centering
    \begin{minipage}[b]{0.48\textwidth}
        \centering
            \caption{\textbf{Ablation study on Bonn RGB-D dataset.} \textbf{Bold} is the best result.}
        \small
        \renewcommand{\arraystretch}{1.1}
        \setlength{\tabcolsep}{1.0mm}
        \begin{tabular}{ccccc}
        \hline
            ~ & ATE & PSNR & Model  Size & Sim. \\ \hline
            w/o Time. & 3.37 & 26.6 & 7.04 & 78.6 \\ 
            w/o Seg. & 3.46 & 27.1 & 7.03 & 78.5\\ 
            w/o Prob. & 3.57 & 27.74 & 22.92 & 79.2 \\ 
            UP-SLAM(Our) & \textbf{3.2} & \textbf{28} & \textbf{7.01}& \textbf{79.5} \\ \hline
        \end{tabular}
        
        \label{tab:ablation}
    \end{minipage}%
    \hfill
    \begin{minipage}[b]{0.5\textwidth}
        \centering
        \caption{\textbf{Tracking results on Scannet dataset.}  \textbf{Bold} is the best result. (ATE RMSE$\downarrow$ [cm])}
        \small
        \renewcommand{\arraystretch}{1.1}
        \setlength{\tabcolsep}{1.0mm}
        \begin{tabular}{c|ccccc|c}
        \hline
            Method & 00 & 59 & 106 & 169 & 207 & Avg. \\ \hline
            Co-SLAM & \textbf{7.1} & 11.1 & 9.4 & \textbf{5.9} & 7.1 & 8.8 \\ 
            SplaTAM & 12.8 & 10.1 & 17.7 & 12.1 & 7.5 & 11.9 \\ 
            DG-SLAM & 7.9 & 11.5 & \textbf{8.0} & 8.3 & 8.2 & 8.6 \\ \hline
            UP-SLAM(Our) & 8.2 & \textbf{7.3} & 8.2 & 8.8 & \textbf{7.0} & \textbf{7.9} \\ \hline
        \end{tabular}
        
        \label{tab:ate_scannet}
    \end{minipage}
\end{table}

\begin{table}[t]
\renewcommand{\arraystretch}{1.0}
 \setlength{\tabcolsep}{0.7mm}
	\centering
        \caption{\textbf{Rendering performance comparison of SLAM methods on Bonn RGB-D dataset.}}
    \small
	\begin{tabular}{c|ccccccc|c}
		\hline
Sequence & Metric & Balloon  & Balloon2  & Ball\_track & Ps\_track   & Ps\_track2   & movbox2 & Avg.  \\ \hline
\multirow{3}{*}{SplaTAM}
& PSNR$\uparrow$   & 20.55 & 18.74 & 20.44 & 17.41 & 16.27 & 22.43 & 19.30       \\
& SSIM$\uparrow$   & 0.829  & 0.756  & 0.819  & 0.438  & 0.625  & 0.881 & 0.724       \\
& LPIPS$\downarrow$  & 0.184  & 0.247  & 0.207  & 0.307  & 0.339  & 0.158 & 0.240    \\ \hline
\multirow{3}{*}{Photo-SLAM} 
& PSNR$\uparrow$   & 20.82 & 22.80 & \underline{25.31} & 22.63 & \underline{23.72} & \underline{25.60} & \underline{23.48}       \\
& SSIM$\uparrow$   & 0.814  & 0.830  & 0.833  & 0.803  & 0.814  & 0.859 &0.825         \\
& LPIPS$\downarrow$  & 0.210  & 0.175  & 0.183  & 0.272  & 0.254  & 0.159  & 0.208       \\ \hline
\multirow{3}{*}{DG-SLAM}      
& PSNR$\uparrow$   & 17.15 & 16.32 & 16.63 & 18.62 & 17.60 & 18.48  & 17.46    \\
& SSIM$\uparrow$   & 0.779 & 0.752  &  0.672  & 0.748  & 0.715  & 0.805  & 0.745   \\
& LPIPS$\downarrow$  &  0.393  & 0.396 &  0.535  & 0.506  & 0.540  & 0.415  & 0.464    \\ \hline
\multirow{3}{*}{WildGS-SLAM(RGB)}      
& PSNR$\uparrow$   & \underline{25.02} & \underline{24.24} & 22.33 &  \underline{22.93} &  22.82 & 23.25 &23.43    \\
& SSIM$\uparrow$   & \textbf{0.961} & \textbf{0.950} & \textbf{0.929} &  \textbf{0.941} &  \textbf{0.946} & \textbf{ 0.921} &\textbf{0.941}     \\
& LPIPS$\downarrow$  & 0.143 & 0.154 & 0.212 &  0.198 &  0.163 & 0.245   &\underline{0.185}   \\ \hline

\multirow{3}{*}{UP-SLAM(Our)}          
& PSNR$\uparrow$   & \textbf{29.31} & \textbf{28.03} & \textbf{27.58} &\textbf{27.98} & \textbf{27.47}& \textbf{27.67}   & \textbf{28.0}   \\
& SSIM$\uparrow$   & \underline{0.921} & \underline{0.919} & \underline{0.886} & \underline{0.899}  & \underline{0.896} & \underline{0.903}  & \underline{0.904}    \\
& LPIPS$\downarrow$  & \textbf{0.089} & \textbf{0.100} & \textbf{0.144} & \textbf{0.128} & \textbf{0.118} & \textbf{0.128} & \textbf{0.117}    \\ \hline

	\end{tabular}

    \label{tab:render_bonn}
    \vspace{-2mm}
\end{table}
\begin{table}[b]
 \vspace{-4mm}
    \centering
    \caption{\textbf{Additional experiments on Bonn balloon sequence.} SplaTAM and DG-SLAM do not involve refinement, hence no extra time is spent.  "-" indicates the absence of this component. \textbf{Bold} is the best result.}
    \small
    \begin{tabular}{cccc}
    \hline
         ~ & Avg.$/$frame$\downarrow$[ms] & Total Time(+refine)$\downarrow$[s] & Model Size$\downarrow$[MB] \\ \hline
        SplatTAM & 4046 & 1776.54(+0) & 29.9 \\ 
        WildGS-SLAM & 1838 & 1526.584(+719.61) & 8.8\\ 
        DG-SLAM & 1011 & \textbf{444.1684}(+0) & - \\ 
        UP-SLAM(Our) & \textbf{78} & 694.814(+660.309) & \textbf{4.9}\\ \hline
    \end{tabular}
        
\label{tab:addition}
\vspace{-2mm}
\end{table}

\subsection{Evaluation of Tracking Performance}

\paragraph{Dynamic Scenes.}
Our method achieves an average improvement of 59.8\% in localization accuracy compared to DG-SLAM. Notably, as shown in Table~\ref{tab:ate_wild}, it improves average localization accuracy by 84.7\%, primarily because DG-SLAM achieves open-set capability based on historical geometric information, which is less robust in complex dynamic environments. While DynaSLAM performs well in Table~\ref{tab:ate_tum} due to its predefined dynamic object handling strategy, it exhibits noticeable drift in Tables~\ref{tab:ate_bonn},\ref{tab:ate_wild}. This degradation arises from the presence of numerous dynamic objects that are difficult to predefine in those datasets, especially in the \emph{Table2} and the \emph{Umbrella} (Umb.) sequences.

\paragraph{Static Scenes.}
UP-SLAM is evaluated on the public static ScanNet~\cite{scannet} dataset to assess robustness. While dynamic object recognition is utilized to improve the robustness of SLAM systems in dynamic environments, inaccurate recognition can adversely affect localization accuracy in static scenes. As shown in Table~\ref{tab:ate_scannet}, our approach achieves an average improvement of 10.2\% in localization accuracy over SLAM systems designed for static environments. Moreover, it achieves an 8.1\% improvement on average compared to DG-SLAM, which is also designed for dynamic scenes. These results demonstrate that our approach maintains strong performance in both static and dynamic environments.

\subsection{Evaluation of Mapping Performance}
\begin{figure}[t]
    \centering
    \includegraphics[scale=0.22]{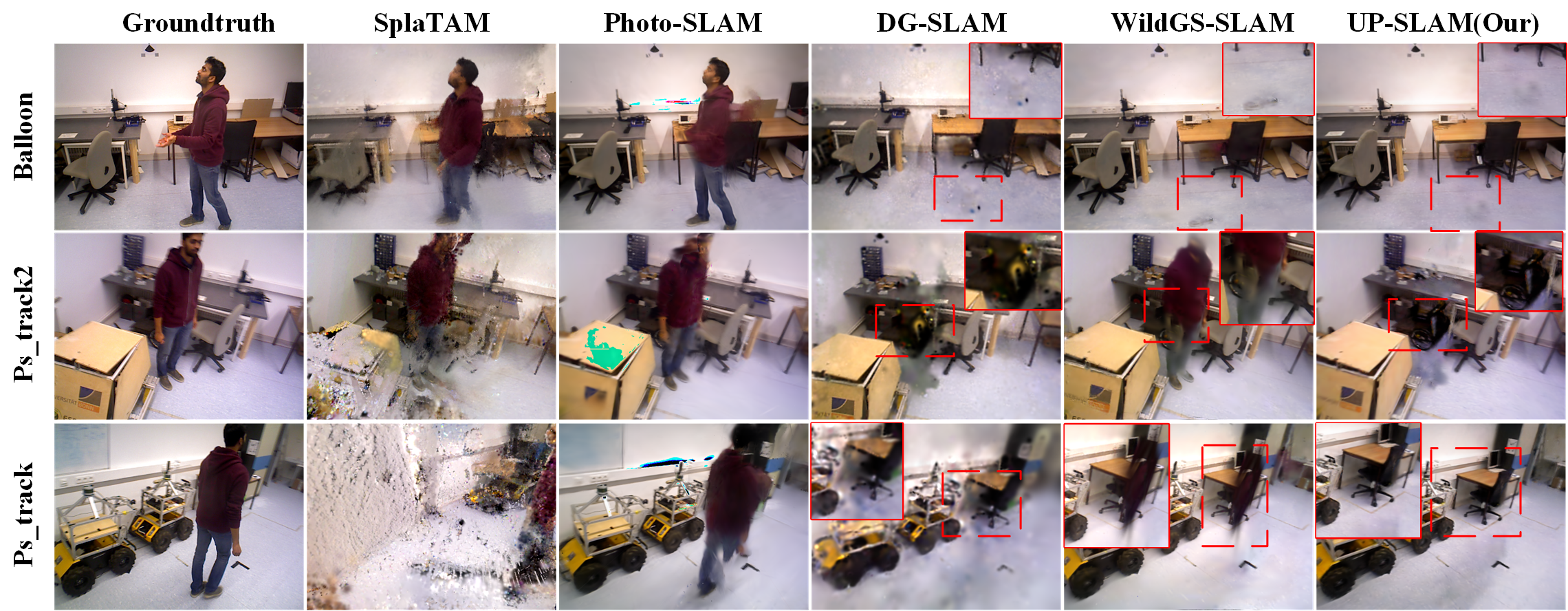}
    \caption{\textbf{The rendering visualization results on Bonn RGB-D dataset.} The red box is a zoom-in of the red dashed box.}
		\label{fig:render_bonn}
        \vspace{-4mm}
\end{figure}

As reported in Table~\ref{tab:render_bonn}, our method achieves a notable improvement in rendering quality, with an average increase of 5.47 dB in PSNR. Photo-SLAM achieves rendering quality comparable to WildGS-SLAM, primarily due to its robustness in low-dynamic sequences (e.g., \emph{Ball\_track} and \emph{Mv\_box2}). However, in highly dynamic environments, localization failures diminish the practical significance of the rendering results. Additionally, the absence of a robust Gaussian primitive initialization strategy in DG-SLAM leads to incomplete reconstructions, significantly degrading rendering quality. Fig.~\ref{fig:render_bonn} provides a visual comparison of the rendered results. The two static SLAM methods, SplaTAM and Photo-SLAM, fail to generate a static map. Both DG-SLAM and the monocular dynamic SLAM method WildGS-SLAM exhibit varying degrees of failure. In contrast, UP-SLAM effectively removes dynamic objects and constructs a high-fidelity, artifact-free static map.

% \begin{figure}[t]
%     \centering
%     \includegraphics[scale=0.26]{figure/wild.png}
%     \caption{\textbf{The rendering visualization results on MoCap.} The red box is a zoom-in of the red dashed box.}
% 		\label{fig:render_wild}
% \end{figure}

\subsection{Ablation Study}
Localization accuracy ($\downarrow$[cm]), rendering PSNR ($\uparrow$[dB]), model size ($\downarrow$[MB]), and DINO feature similarity (Sim. $\uparrow$[\%]) are quantified to comprehensively evaluate the contribution of each component, as summarized in Table~\ref{tab:ablation}. The temporal encoding primarily enhances rendering quality, yielding an average PSNR improvement of 1.4 dB. The segmentation module proves critical for both localization and rendering. In the initialization phase, the presence of non-rigid objects hinders segmentation of entire dynamic regions, causing potentially dynamic keypoints to be incorrectly added as static ones to the map, where they are then used as landmarks for localization. The probabilistic anchor update module significantly reduces model size, improving its suitability for deployment on embedded platforms. Without anchor updates, Gaussian primitives cannot be effectively pruned, leading to slower map updates and weakened residual feedback to the tracking thread, ultimately degrading pose estimation accuracy. Moreover, UP-SLAM improves similarity scores to nearly 80\%, demonstrating its potential for downstream applications such as object-level navigation and semantic understanding.

\subsection{Runtime Analysis}
Table~\ref{tab:addition} presents the runtime analysis. By decoupling motion mask generation from map optimization, our system achieves a parallel tracking and mapping architecture. This enables a processing rate of 12 Hz per frame, meeting the real-time localization requirements for robotics. Compared to WildGS-SLAM, we adopt the same number of refine optimization iterations but achieve a 2× speed-up. DG-SLAM excludes segmentation time from the reported runtime. While it exhibits the fastest overall runtime, our method offers a better balance between reconstruction quality and localization speed. This trade-off is reasonable, given that mapping is typically less constrained by real-time requirements than tracking. Additionally, the use of probabilistic anchor updates and MLP-based Gaussian attribute encoding significantly reduces the overall model size.

{
\newpage
\small
\bibliographystyle{IEEEtran}
\bibliography{neurips_2025}
}
\newpage
\appendix

\section{Technical Appendices}
\label{sec:appendix}
\begin{figure}[b]
    \centering
    \includegraphics[scale=0.26]{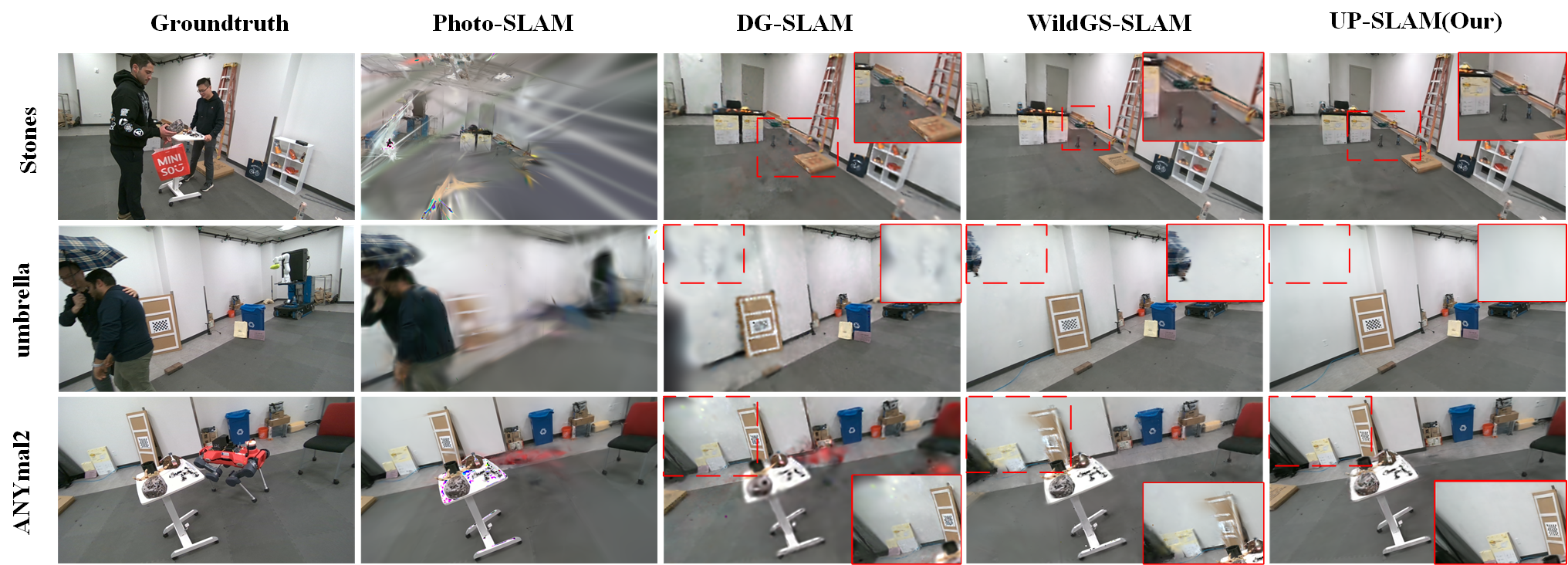}
    \caption{\textbf{The rendering visualization results on MoCap RGB-D dataset.} The red box is a zoom-in of the red dashed box.}
		\label{fig:render_wild}
\end{figure}

\begin{figure}[b]
    \centering
    \includegraphics[scale=0.4]{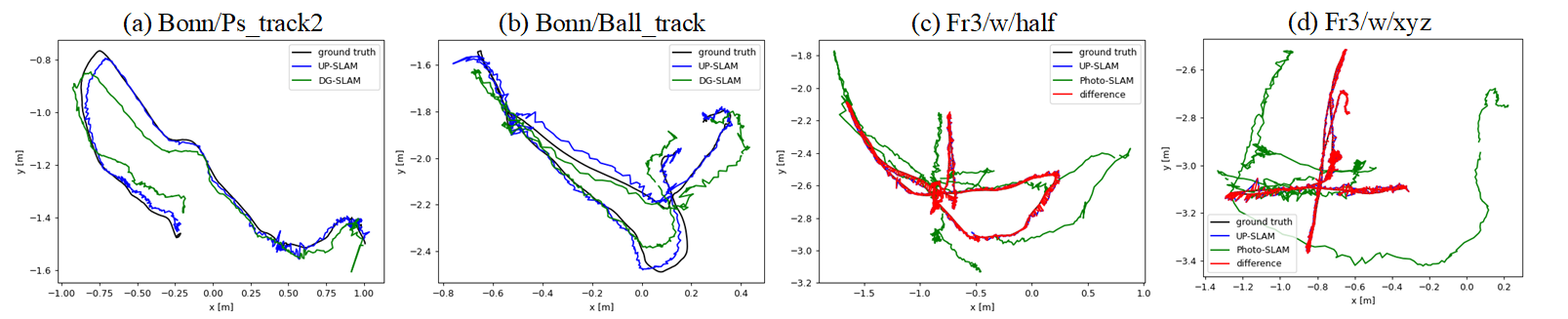}
    \caption{\textbf{Localization trajectory comparisons.}  (a) and (b) show the trajectory comparison with DG-SLAM, while (c) and (d) compare with Photo-SLAM. The red lines indicate the deviation from the ground truth, with shorter lengths reflecting higher localization accuracy. }
		\label{fig:traj}
\end{figure}

\paragraph{Dataset.}
Our method is evaluated on the TUM RGB-D Dataset~\cite{tum}, the Bonn RGB-D Dynamic Dataset~\cite{bonn}, the MoCap RGB-D Dataset~\cite{wildslam}, and the static environment ScanNet Dataset~\cite{scannet}. In the TUM dataset, the primary dynamic objects are humans. In the Bonn dataset, in addition to humans, a small number of other dynamic objects, such as balloons and boxes, are present. The image resolution for both datasets is 640×480. In contrast, the MoCap RGB-D dataset features a higher image resolution of 1280×720 and contains a large number of unstructured dynamic objects beyond humans, such as robotic dogs, rocks, and other irregular entities. This poses significant challenges for dynamic SLAM methods that rely on predefined dynamic object categories. In the selected ScanNet dataset sequences, RGB-D streams with a resolution of 640×480 are captured in real indoor static environments, and the ground-truth camera poses are provided.

\paragraph{Baseline.}
The baselines we selected are the same as those in Sec.~\ref{sec:exp}. For better analysis, we group methods of the same category together.

\paragraph{Metrics.}
For tracking evaluation, the official TUM evaluation script is used to jointly align the estimated and ground-truth poses and compute the RMSE of the Absolute Trajectory Error (ATE). In dynamic scenes, traditional rendering-based evaluation metrics (e.g., PSNR, SSIM, LPIPS) may become unreliable, as they are designed to assess performance under static conditions. This misalignment contrasts with our goal of reconstructing a clean static background despite the presence of dynamic elements. To address this, we adopt pre-generated dynamic object masks obtained using Ground-DINO~\cite{groundingdino} on the Bonn dataset. During evaluation, these masks are applied to exclude dynamic regions, ensuring that the rendering quality metrics reflect the fidelity of the static background reconstruction. Moreover, we will release the generated mask dataset to facilitate future research.

\begin{table}[t]
    \renewcommand{\arraystretch}{1.0}
    \setlength{\tabcolsep}{1.2mm}
    \centering
    \caption{\textbf{Tracking results on Bonn RGB-D dataset (ATE RMSE$\downarrow$ [cm]).}  "-" indicates unavailable data because the related work is not open. \textbf{Bold} is the best result, and \underline{underline} is the second best result.}
    \begin{tabular}{c|cccccc|c}
    \hline
        Method & Balloon & Balloon2 & Ball\_track & Ps\_track & Ps\_track2 & Mv\_box2 & Avg. \\ \hline
        ORB-SLAM3 & 5.8 & 17.7 & \underline{3.1} & 70.7 & 77.9 & \underline{3.5} & 29.78 \\ \hline
        ReFusion & 17.5 & 25.4 & 30.2 & 28.9 & 46.3 & 17.9 & 27.7 \\ 
        DynaSLAM & 3 & \underline{2.9} & 4.9 & 6.1 & 7.8 & 3.9 & \underline{4.76} \\ \hline
        iMap & 14.9 & 67 & 24.8 & 28.3 & 52.8 & 28.3 & 36.01 \\ 
        NICE-SLAM & X & 66.8 & 21.2 & 54.9 & 45.3 & 31.9 & 36.68 \\ 
        Vox-Fusion & 65.7 & 82.1 & 43.9 & 128.6 & 162.2 & 31.9 & 85.73 \\ 
        Co-SLAM & 28.8 & 20.6 & 38.3 & 61 & 59.1 & 70 & 46.3 \\ 
        ESLAM & 22.6 & 36.2 & 12.4 & 48 & 51.4 & 17.7 & 31.38 \\ \hline
        RoDyn-SLAM & 7.9 & 11.5 & 13.3 & 14.5 & 13.8 & 12.6 & 12.26 \\ \hline
        Photo-SLAM & 6.9 & 26 & 3.2 & 76.4 & 87.4 & 3.6 & 33.91 \\ 
        GS-SLAM & 37.5 & 26.8 & 31.9 & 46.8 & 50.4 & 4.8 & 33.03 \\ 
        SplaTAM & 36.12 & 35.1 & 12.93 & 128.7 & 136.5 & 20.6 & 61.65 \\ \hline
        DG-SLAM & 3.7 & 4.1 & 10 & \underline{4.5} & \underline{6.9} & 3.5 & 5.45 \\ 
        GassiDy & \textbf{2.6} & 7.6 & - & 10.3 & 13 & 5.4 & 7.78 \\ \hline
        UP-SLAM(Our) &\underline{2.8} & \textbf{2.7} & \textbf{2.9} & \textbf{4.0} & \textbf{3.6} & \textbf{3.2} & \textbf{3.2} \\ \hline
    \end{tabular}
    
\label{tab:ate_bonn_a}
% \vspace{-4mm}

\end{table}
\begin{table}[t]
    \renewcommand{\arraystretch}{1.1}
    \setlength{\tabcolsep}{0.3mm}
    \centering
    \caption{\textbf{Tracking results on MoCap RGB-D dataset (ATE RMSE$\downarrow$ [cm]).} "-" indicates unavailable data because the related work is not open. "X" denotes a tracking failure. \textbf{Bold} is the best result, and \underline{underline} is the second best result.}
    \begin{tabular}{c|cccccccccc|c}
    \hline
        Method & ANY1 & ANY2 & Ball & Crowd & Person & Racket & Stones & Table1 & Table2 & Umb. & Avg. \\ \hline
        ReFusion & 4.2 & 5.6 & 5 & 91.9 & 5 & 10.4 & 39.4 & 99.1 & 101 & 10.7 & 37.23 \\ 
        DynaSLAM & 1.6 & \textbf{0.5} & \textbf{0.5} & 1.7 & \textbf{0.5} & \textbf{0.8} & 2.1 & \underline{1.2} & \underline{34.8} & 34.7 & 7.84 \\ \hline
        
        NICE-SLAM & X & 123.6 & 21.1 & X & 150.2 & X & 134.4 & 138.4 & X & 23.8 & - \\ \hline
        Photo-SLAM & 79.5 & 11.8 & 50.3 & 105.9 & 27.5 & 38.23 & 113.5 & 39.1 & 64.8 & 84 & 61.46 \\\hline
         
        DG-SLAM & \underline{1.2} & 2.1 & 0.8 & \underline{1.3} & 1.5 & 1.6 & \underline{1.5} & 2 & 57.9 & \underline{1.35} & \underline{7.06} \\ \hline
        UP-SLAM(Our) & \textbf{0.4} & \underline{0.6} & \underline{0.6} & \textbf{1.1} & \underline{1.1} & \underline{0.9} & \textbf{1.0} & \textbf{0.7} & \textbf{3.6} & \textbf{0.8} & \textbf{1.08} \\ \hline
    \end{tabular}
    
\label{tab:ate_wild_a}

\end{table}
% \usepackage[normalem]{ulem}
\begin{table}[t]
\renewcommand{\arraystretch}{1.1}
\setlength{\tabcolsep}{1.2mm}
\centering
\caption{\textbf{Tracking results on TUM RGB-D dataset (ATE RMSE$\downarrow$ [cm]).} "-" indicates unavailable data because the related work is not open. "X" denotes a tracking failure. \textbf{Bold} is the best result, and \underline{underline} is the second best result.}
\begin{tabular}{c|ccccc|c}
\hline
Method     & Fr3/w/xyz & Fr3/w/half & Fr3/w/static & Fr3/s/xyz & Fr2/desk\_person & Avg.          \\ \hline
ORB-SLAM3  & 28.1      & 30.5       & 2.0          & 1.0       & 1.5              & 12.62         \\ \hline
ReFusion   & 9.9       & 10.4       & 1.7          & 4.0       & -                & 6.5           \\
DynaSLAM   & \textbf{1.5}  & 2.9        & 0.7          & 1.6       & \underline{0.9}              & 1.52    \\
EM-Fusion  & 6.6       & 5.1        & 1.4          & 3.7       & -                & 4.2           \\ \hline
iMap & 111.5 & X & 137.3 & 23.6 & 119 & 97.85 \\ 
NICE-SLAM  & 113.8     & X          & 88.2         & 7.9       & X                & 69.96         \\
Vox-Fusion & 146.6     & X          & 109.9        & 3.8       & X                & 86.76         \\
Co-SLAM & 51.8 & 105.1 & 49.5 & 6 & 7.6 & 44 \\ 
ESLAM      & 45.7      & 60.8       & 93.6         & 7.6       & X                & 51.92         \\ \hline
RoDyn-SLAM & 8.3       & 5.6        & 1.7          & 5.1       & 5.6              & 5.26          \\ \hline
Photo-SLAM & 60.4      & 35.7       & 13.7         & \underline{1.0} & \textbf{0.6}              & 22.28         \\
GS-SLAM    & 37.2      & 60.0       & 8.4          & 2.7       & 8.6              & 23.38         \\
SplaTAM    & 140.6     & 153.58     & 90.36        & 1.6       & 5.4              & 78.30         \\ \hline
DG-SLAM    & 1.7       & \textbf{1.8}  & \underline{0.7}  & 1.0       & 3.2              & 1.68          \\
GassiDy    & 3.5       & 3.7        & \textbf{0.6} & -         & -                & \underline{2.6}           \\ \hline
UP-SLAM(Our) & \underline{1.6} & \underline{2.6}  & \underline{0.7}    & \textbf{0.9}  & 1.3              & \textbf{1.42} \\\hline
\end{tabular}

\label{tab:ate_tum_a}
\end{table}

\begin{table}[t]
    \centering
    \caption{\textbf{Tracking results on Scannet dataset (ATE RMSE$\downarrow$ [cm]).}  \textbf{Bold} is the best result.}
    \setlength{\tabcolsep}{1.7mm}
    \renewcommand{\arraystretch}{1.0}
    \begin{tabular}{c|ccccc|c}
    \hline
        Method & 00 & 59 & 106 & 169 & 207 & Avg. \\ \hline
        NICE-SLAM & 12 & 14 & \textbf{7.9} & 10.9 & \textbf{6.2} & 10.7 \\ 
        Co-SLAM & \textbf{7.1} & 11.1 & 9.4 & \textbf{5.9} & 7.1 & 8.8 \\ 
        Point-SLAM & 10.2 & 7.8 & 8.7 & 22.2 & 14.8 & 12.2 \\ 
        Vox-Fusion & 68.8 & 24.1 & 8.4 & 27.2 & 9.4 & 27.58 \\ \hline
        GS-SLAM & 13.6 & 7.6 & 8.1 & 13.7 & 34.6 & 15.1 \\ 
        SplaTAM & 12.8 & 10.1 & 17.7 & 12.1 & 7.5 & 11.9 \\ \hline
        DG-SLAM & 7.9 & 11.5 & 8.0 & 8.3 & 8.2 & 8.6 \\ 
        UP-SLAM(Our) & 8.2 & \textbf{7.3} & 8.2 & 8.8 & 7.0 & \textbf{7.9} \\ \hline
    \end{tabular}
    
    \label{tab:ate_scannet_a}
\end{table}

\paragraph{Implementation Details.}
 UP-SLAM is fully implemented in C++ and CUDA, and runs on a desktop equipped with Intel i7-12700KF and an NVIDIA RTX 4060ti 16G GPU. We set the loss weight: \{$\lambda$,$\lambda_{1}$,$\lambda_{2}$,$\lambda_{3}$,$\lambda_{4}$,$\lambda_{5}$\}=\{$0.8$,$0.6$,$1.0$,$0.4$,$0.01$\}, the residuals weight is \{$\lambda_{1}^t$,$\lambda_{2}^t$,$\lambda_{3}^t$\}=\{$0.25$,$0.7$,$0.1$\}, the refinement iteration count: 20000, and low-dimensional $N_{l}=16$.

\paragraph{MLP Implementation Details.}
The architecture of MLP $\mathcal{F}_d$ is: $\mathrm{LINER}\to\mathrm{SoftPlus}\to\mathrm{LINER}$. The hidden layer has 32 dimensions. 

The architecture of MLP $\mathcal{F}_m$ is: $\mathrm{Convolution}\to\mathrm{ReLU}\to\mathrm{Convolution}$. The hidden layer has 128 dimensions. 

The architecture of MLP $\mathcal{F}_u$ is: $\mathrm{Convolution}\to\mathrm{ReLU}\to\mathrm{Convolution}\to\mathrm{SoftPlus}$. The hidden layer has 128 dimensions.

For other network architectures $\mathcal{F}_c$,$\mathcal{F}_a$, $\mathcal{F}_s$, $\mathcal{F}_q$ please refer to~\cite{scaffold}.

\paragraph{Tracking Experiments.}
In Tables~\ref{tab:ate_bonn_a}, ~\ref{tab:ate_wild_a}, ~\ref{tab:ate_scannet_a}, we additionally compare several well-known algorithms to enable a more comprehensive analysis and comparison from multiple perspectives. In Fig.~\ref{fig:traj}, representative trajectories are presented for reference and comparative analysis.

\paragraph{Mapping Experiments.}
In Fig.~\ref{fig:render_wild}, for the stones sequence, both DG-SLAM and WildGS-SLAM produce red artifacts, whereas UP-SLAM successfully renders a clean background. In the umbrella sequence, although DG-SLAM reconstructs a static background, the resulting map exhibits numerous holes.

\paragraph{Limitations.}
While UP-SLAM demonstrates strong performance in dynamic environments through parallel tracking and uncertainty-aware dynamic object filtering, several limitations persist. To support model compression and automatic Gaussian management, we introduce probabilistic anchors, from which Gaussian attributes are decoded via shallow MLPs. However, this decoding process increases optimization time and may introduce noise into the residual computations, particularly under limited training iterations.

% \paragraph{Acknowledgments and Disclosure of Funding.}
% This research was supported by the National Natural Science Foundation of China (Grant 62373329), the Baima Lake Laboratory Joint Funds of the Zhejiang Provincial Natural Science Foundation of China (Grant No. LBMHD24F030002) and the Natural Science Foundation of Zhejiang Province of China (Grant No.Z25F030018).

\end{document}